\documentclass[letterpaper, 10 pt, conference]{ieeeconf}  

\IEEEoverridecommandlockouts                              

\usepackage{amsmath} 
\usepackage{amssymb}  
\usepackage{mathtools}

\usepackage{balance}
\usepackage{hyperref}

\usepackage{booktabs}  
\usepackage{longtable} 
\usepackage{multirow}  
\usepackage{graphicx}  
\usepackage{siunitx}  

\title{\LARGE \bf

Data-augmented Learning of Geodesic Distances in Irregular Domains through Soner Boundary Conditions
}

\author{Rafael I. Cabral Muchacho and Florian T. Pokorny%
\thanks{This work was partially supported by the Wallenberg AI, Autonomous Systems and Software Program (WASP) funded by the Knut and Alice Wallenberg Foundation.} 
\thanks{The authors are with RPL, EECS, KTH Royal Institute of Technology, Stockholm, Sweden { \tt\small \{ricm, fpokorny\}@kth.se}}
}

\begin{document}

\maketitle
\thispagestyle{empty}
\pagestyle{empty}

\begin{abstract}
Geodesic distances play a fundamental role in robotics, as they efficiently encode global geometric information of the domain.
Recent methods use neural networks to approximate geodesic distances by solving the Eikonal equation through physics-informed approaches. 
While effective, these approaches often suffer from unstable convergence during training in complex environments. 
We propose a framework to learn geodesic distances in irregular domains by using the Soner boundary condition, and systematically evaluate the impact of data losses on training stability and solution accuracy.
Our experiments demonstrate that incorporating data losses significantly improves convergence robustness, reducing training instabilities and sensitivity to initialization. These findings suggest that hybrid data-physics approaches can effectively enhance the reliability of learning-based geodesic distance solvers with sparse data.
\end{abstract}

\section{Introduction}


Global structure plays a key role in motion planning and control, complementing local reactive methods to ensure optimality and feasibility. Partial differential equations (PDEs) provide a principled way to encode global geometric information through local quantities. 
In particular, the Eikonal equation models shortest-path distances within a domain, making it a fundamental tool for computing geodesic distances in planning and navigation.

Geodesic distances in bounded domains are directly related to path planning and provide optimal path information while satisfying domain constraints. While geodesic distances have analytical solutions in simple domains, computing them in arbitrary domains typically requires numerical methods or domain-specific knowledge.

Classical numerical solvers such as the Fast Marching Method (FMM) and Fast Sweeping Method (FSM) provide accurate solutions but scale poorly in high-dimensional spaces. Neural approaches offer an alternative, bypassing explicit discretization, but often struggle with convergence stability~\cite{sethian1996fast, zhao2005fast, smith2020eikonet, grubas2023neural}.

This work focuses on improving the stability of neural solvers for geodesic distances. While physics-informed losses can enforce the Eikonal equation, neural training remains sensitive to initialization and optimization challenges. 
We examine how data supervision enhances convergence. 
Based on findings from physics-informed neural networks (PINNs), we hypothesize that incorporating even sparse data improves training stability and accuracy compared to physics-only losses~\cite{cuomo2022scientific}.

To ensure a well-posed formulation, we decouple domain boundaries from speed models by encoding boundaries through the Soner condition~\cite{deckelnick2011numerical, soner1986optimal}, which allows for exact solutions of the geodesic distance.
For path planning this brings the additional benefit of separating safety constraints from spatial preferences in the domain.

We systematically evaluate the role of data losses through controlled ablation studies on data \textit{quantity} and \textit{quality}. 
Data quantity is represented by the size of the available labeled dataset, and quality is represented by the noise level used to corrupt the dataset.
Our findings suggest that sparse and noisy data significantly improves convergence, reducing reliance on extensive supervision while maintaining accuracy. 

Specifically,
\begin{enumerate}
    \item We propose a physics-informed learning framework to approximate geodesic distances in irregular domains and integrates data losses for improved convergence.
    \item We perform a systematic study on the effect of data supervision in neural geodesic solvers, analyzing both data quantity and quality.
    \item We provide empirical evidence that sparse but well-placed data points significantly enhance convergence, reducing the need for large datasets.
\end{enumerate}

These results highlight the potential of hybrid physics-data approaches in neural motion planning, bridging classical PDE-based solvers with imitation learning-based methods.

\begin{figure}
    \centering
    \includegraphics[width=\linewidth]{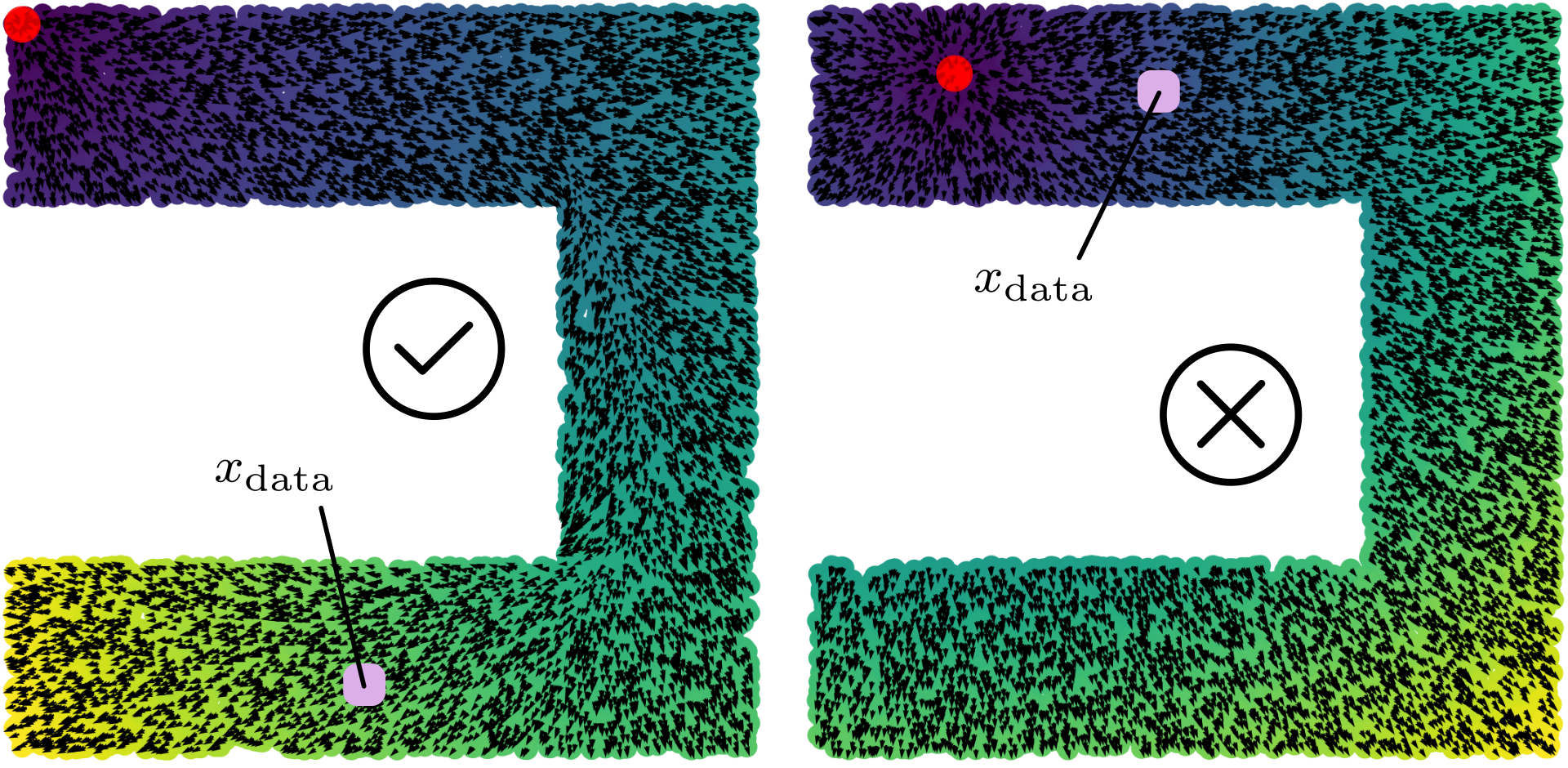}
    \caption{This figure demonstrates that even sparse data can significantly influence the learned geodesic distance functions. The color map represents the estimated geodesic distance to the source or goal point (red), and the arrows depict the gradient field. The left and right figures correspond to different placements of a single supervision point $x_{\mathrm{data}}$. While the right case converges to a poor minimum, the left case approaches the correct solution.}
    \label{fig:one-matters}
\end{figure}

\section{Background}

\subsection{Lengths and Distances}

In a bounded domain \(\Omega \subset \mathbb{R}^n\), the distance between points is defined in terms of the Euclidean length of a path constrained to the domain. Given a differentiable path \(\gamma: [0,1] \to \Omega\), its length is given by
\begin{align}
    \mathrm{length}(\gamma) = \int_0^1 \|\gamma^\prime(t)\| \, dt.
\end{align}

The geodesic distance between two points \(x, y \in \Omega\) is the shortest path length within the domain
\begin{align}
    d(x, y) = \inf_{\gamma} \mathrm{length}(\gamma),
\end{align}
where the infimum is taken over all differentiable paths \(\gamma\) such that \(\gamma(0) = x\) and \(\gamma(1) = y\).
A related concept is the distance-to-goal function, which assigns each point its shortest distance to a target set \(\Omega_g \subset \Omega\)
\begin{align}
    d(x) = \min_{y \in \Omega_g} d(x, y).
\end{align}
This formulation is commonly used in motion planning and front propagation methods.
The geodesic distance function satisfies the Eikonal equation
\begin{align}
    \|\nabla d(x)\| = c(x) \quad x \in \Omega,
\end{align}
with a speed model $c(x)$, and subject to boundary conditions \(d(x) = 0\) for \(x \in \Omega_g\)~\cite{zhao2005fast}. Since the gradient \(\nabla d\) may be discontinuous, the solution is understood in the viscosity sense, ensuring well-posedness and uniqueness in irregular domains~\cite{deckelnick2011numerical}.

\subsection{Solving the Eikonal Equation}

Computing geodesic distances in bounded domains is equivalent to solving the Eikonal equation. 
Traditional numerical approaches and neural network-based methods provide two distinct paradigms, each with its own trade-offs.


Classical numerical solvers, such as the Fast Marching Method (FMM) and Fast Sweeping Method (FSM), discretize the domain and solve for the geodesic distance using grid-based schemes. 
These methods are (i) efficient in low dimensions due to structured grids and ordered updates, (ii) highly accurate when properly discretized, benefiting from well-studied convergence properties, and (iii) limited by grid resolution, making them computationally expensive in higher dimensions~\cite{sethian1996fast, treister2016fast, belyaev2015variational, crane2013geodesics}.


Physics-informed neural networks (PINNs) and related deep learning models offer an alternative by approximating the geodesic distance through neural networks \cite{bin2021pinneik, smith2020eikonet, grubas2023neural}. 
Unlike grid-based solvers, neural methods (i) scale more flexibly to higher dimensions, bypassing explicit discretization, (ii) provide smooth and differentiable solutions, useful in gradient-based planning, (iii) can require extensive training data or physics-informed losses to generalize accurately.

\subsection{Neural Eikonal Methods in Robotics}

Recent advances on neural time fields (NTField)~\cite{ni2022ntfields} have led to interesting and promising Eikonal-based methods in motion planning and mapping~\cite{ni2023progressive, liu2024physics}, including methods in non-euclidean geometry~\cite{li2024riemannian}.
NTField methods aim to generate motion plans by solving an Eikonal equation in the configuration space of a given robot, computing shortest paths by integrating along the gradient of the solution.

The method, as parametrized in~\cite{liu2024physics}, can be summarized through the equations
\begin{align}
    \frac{1}{S(x_2)} &= \lVert \nabla_{x_2}T(x_1, x_2)\rVert \label{eq:speedmodel-Eik}\\
    T(x_1, x_2) &= \log(\tau_\theta(x_1, x_2))^2\lVert x_1 - x_2\rVert, \label{eq:speedmodel-T}
\end{align}
with  
\begin{align}
    S^*(x) = \frac{s_\mathrm{const}}{\mathbf{d}_\mathrm{max}}\times \mathrm{clip}\left(  \mathbf{d}(x, \mathcal{X}_\mathrm{obs}), \mathbf{d}_\mathrm{min}, \mathbf{d}_\mathrm{max}  \right), \label{eq:speedmodel-S}
\end{align}
where $\mathbf{d}$ represents a distance function to obstacles $\mathcal{X}_\mathrm{obs}$, $\tau_\theta$ is a scalar parametric function, and $S^*$ is the ground truth speed model that approximately encodes the domain boundary at $\mathbf{d}(x)=0$.
While NTField methods are based on an Eikonal equation,
the focus is on the direct usage of solutions for navigation and planning, rather than on the exact computation of geodesic distances.

Reliable training stability and convergence of geodesic distance approximations through PINNs remains elusive, especially in irregular and complex domains.
Motivated by this observation, we define our research problem and hypothesis.

\section{Problem Statement and Hypothesis}
We consider the approximation of geodesic distances in irregular, bounded domains using PINNs, given a compact domain with well-defined boundary normals.
The aim of this study is to contextualize and empirically assess the effect of data losses on the learning accuracy and stability, and thus inform future design decisions in physics-informed neural motion planning.

To systematically evaluate this, we first introduce a framework that decouples boundary conditions from the speed model. 
Rather than coupling these directly, as done in NTFields~\cite{ni2022ntfields}, we encode domain boundaries through the Soner boundary condition, which is independent of speed model and of the underlying geometry.
This framework allows for exact convergence to geodesic distances, which we use to evaluate of how data losses influence convergence behavior and solution quality.

We expect that the accuracy on geodesic distance estimation achieved with infinite data can be matched by a combination of sparse data and physics-based losses, particularly when enforcing both the Eikonal and Soner boundary conditions. 
If this holds, it underscores the effectiveness of physics-informed learning in reducing data dependence while maintaining reliable geodesic estimation in complex domains.

\section{Framework} \label{sec:framework}

Given a dataset of shortest paths in a static environment, we combine the approaches of imitation learning and PDE-based solvers by using neural networks as a surrogate, and defining a loss function that merges both perspectives in a non-conflicting manner. 

Intuitively, the data-driven regression loss should guide the solver, and the PDE losses can refine and extrapolate the data to unseen domain regions of interest.
Note that the extrapolation or solution refinement does not necessarily need to be to the entire domain; it could be restricted to regions of interest to reduce computation, e.g., tubes around demonstrations or balls around data points.


\subsection{Eikonal Equation in Irregular Domains}

For the physics-based components we build on NTFields~\cite{ni2022ntfields}, by formalizing the constant speed Eikonal equation with gradient-based boundary inequalities to encode domain boundaries, effectively freeing the distance dependent speed model.

The distance function $d: \Omega \to \mathbb{R}_{\geq 0}$ from the goal set $\mathcal{G}$ in the domain $\Omega \subset \mathbb{R}^n$ with domain boundary $\partial \Omega$ is described by the solution to the boundary value problem
\begin{align}
    \lVert \nabla d(x) \rVert = 1 \quad & x \in \Omega \\
    d(x)>0 \quad & x \in \Omega \setminus \mathcal{G} \\
    d(x) = 0 \quad & x \in \mathcal{G} \\
    \nabla d(x) \cdot n(x) \leq 0 \quad & x \in \partial\Omega, \label{eq:soner-condition}
\end{align}
where $n: \partial\Omega \to \mathbb{R}^n$ describes the (inward) unit vector normal to the boundary $\partial \Omega$ at the point $x$. 

In fact, the above equation is a special case of optimal control with state constraints and has a unique solution in the viscosity sense~\cite{deckelnick2011numerical, hahn2024laplacian}.
Its uniqueness implies that it is a \textit{necessary and sufficient} condition in the specification of geodesic distances.

\subsection{Physics Loss}

We implement the Eikonal constraint and the Soner boundary condition as soft constraints through the loss functions 
\begin{align}
    \mathcal{L}_\mathrm{Eik} = (\lVert\nabla d\rVert - 1)^2 \quad & x \in \Omega \\
    \mathcal{L}_\mathrm{Soner} = \mathrm{ReLU}\left(\nabla d \cdot n \right)  \quad & x \in \partial\Omega,
\end{align}
fixing the speed model to $c(x)=1$.
The remaining constraints in the bounded Eikonal are implemented as hard constraints and discussed in Section~\ref{sec:architecture-hard}.
As the Soner loss requires points on the boundary and the corresponding boundary normals, implicitly defined boundaries of robotic manipulators can be sampled as proposed in~\cite{inproceedings}.

\subsection{Data Loss}

For the supervised or data-driven component we use a dataset of shortest paths, or more generally, a dataset consisting of tuples $\left(x_i, d_\mathrm{gt}(x_i), \nabla d_\mathrm{gt}(x_i)\right)$ for points $x_i$ in the domain.
Here $d_\mathrm{gt}(x_i)$ denotes the ground truth geodesic distance between a point $x_i$ and a pre-specified goal.

We define the loss function for value and gradient supervision through the squared norms
\begin{align}
    \mathcal{L}_\mathrm{Data} = \lVert d_\mathrm{gt} - d \rVert^2 + \lVert \nabla d_\mathrm{gt} - \nabla d \rVert^2.
\end{align}
In this formulation we omit explicit weights, as throughout the evaluation we assume constant and equal weights for all losses.

\subsection{Architecture and Hard Constraints}\label{sec:architecture-hard}

We approximate the geodesic distance using a Residual MLP (ResMLP) architecture (5 layers with 128 hidden units per layer, i.e., $5 \times 128$) with hidden \textsc{GELU} activations and a \textsc{SoftPlus} output layer.
The symmetry of the geodesic distance is enforced by averaging the outputs for the concatenated inputs $[x,y]$ and $[y,x]$, resulting in the formulation~\cite{kelshaw2024computing}~\cite{li2024riemannian}
\begin{align}
    d(x,y) &= \| x - y \| \left( 1 + \frac{1}{2} \Bigl( f([x,y]) + f([y,x]) \Bigr) \right), \\[8pt]
    f(z) &= \mathrm{SoftPlus}(\mathrm{ResMLP}_{\mathrm{GELU},\,5\times128}(z)).
\end{align}

The activation functions are chosen to be smooth and continuously differentiable to avoid explicit regularization terms in the physics-based loss function~\cite{implicitLipman}.

The zero-equality and non-negativity conditions of the function are enforced by construction. 
The zero-equality constraint at the source is ensured by using the euclidean distance in free space as a factor.
The parametric factor of the neural network is essentially a coefficient larger than $1$, encoding the non-negativity condition outside of the goal.
The non-negativity condition arises from geodesic distances being larger or equal than the distance in free space, assuming euclidean geometry.

\section{Framework Validation}

In this section we validate our framework and the neural Soner boundary condition.
To this end, we aim to (i) approximate the geodesic distance to a fixed point in a simple domain, (ii) verify the framework's ability to learn the exact geodesic distance, and (iii) highlight the main differences compared to NTField methods~\cite{ni2022ntfields, liu2024physics}.

\begin{figure}
    \centering
    \includegraphics[width=\linewidth]{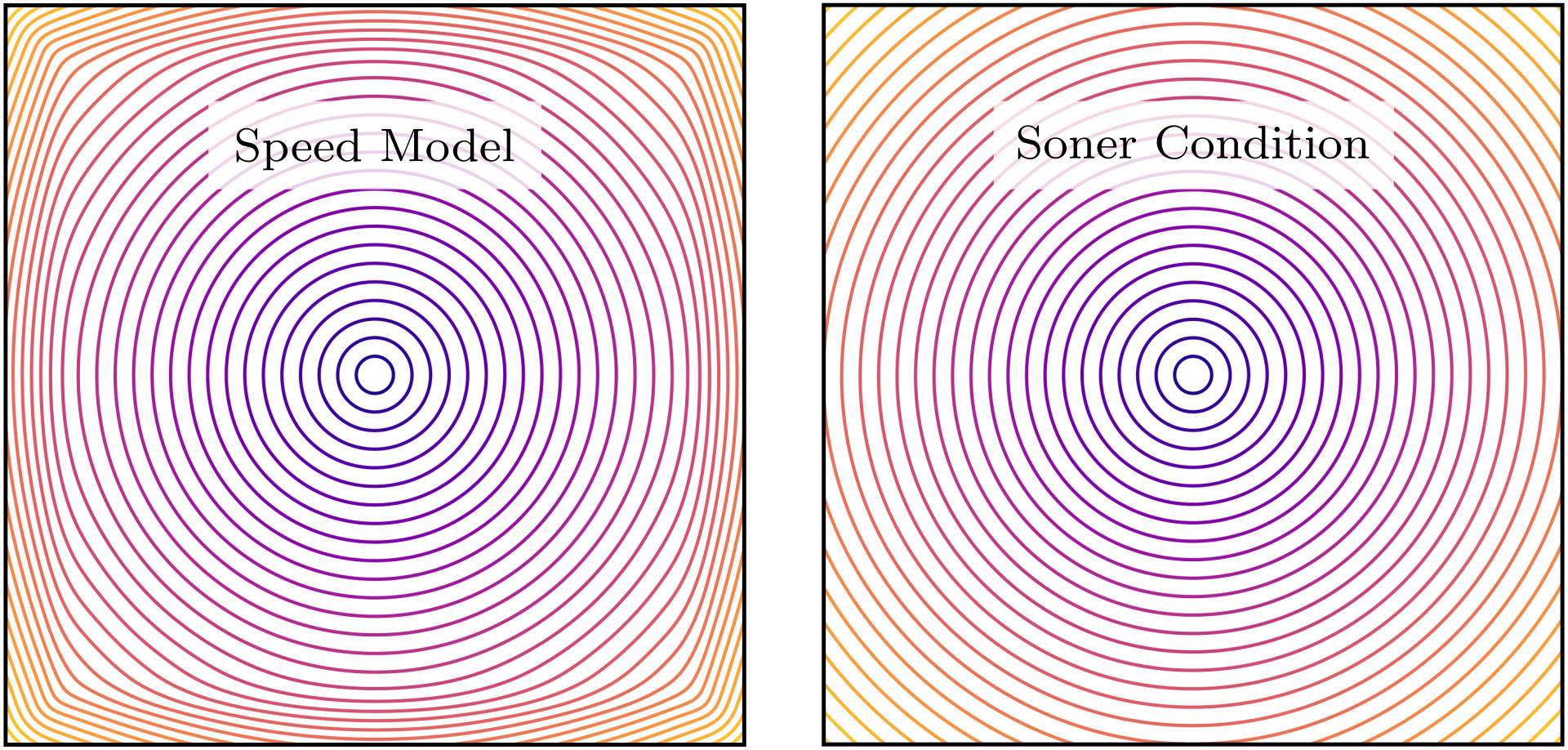}
    \caption{The geodesic distance in flat euclidean space is computed in a square domain of side length $2$, and visualized through equidistant contour lines. The results arising from encoding boundaries through the speed model (left) and through the Soner condition (right) are illustrated in this figure. Note that the speed model variant "squishes" the distance close to the boundary, whereas the Soner condition allows for the exact solution.}
    \label{fig:boundary-comparison}
\end{figure}

The domain is chosen to be a square of side length $2$, centered at the origin. This simple shape is chosen to focus the analysis on the boundary conditions and avoid shape-driven complexities.
We implement an NTField approach described in~\eqref{eq:speedmodel-T} parametrized with
\begin{align}
    \mathbf{d}_\mathrm{min} = 0.1,\quad \mathbf{d}_\mathrm{max} = 0.2,\quad s_\mathrm{const} = 1,
\end{align}
and train the model until convergence.
Illustrations of the resulting landscapes are shown in Figure~\ref{fig:boundary-comparison} through contour plots.

The convex case is accurately approximated, as shown in the square domain plots in Figure~\ref{fig:boundary-comparison}.
The visualization can be used to highlight the differences between the methods. 
It shows that the exact geodesic distance \textit{can} be learned through the framework leveraging the Soner boundary condition, up to limitations in model representation capabilities.

The solution obtained by the NTField method deviates from the geodesic distance close to the boundary, as it is encoded through the speed model.
This is advantageous to the NTField approach when directly used for motion planning through gradient following methods. 
Encoding the boundary through the speed model induces a boundary-avoiding heuristic, and appears to improve the learning stability of physics-only strategies.
In contrast to the approximation of geodesic distances through speed-model based methods, using the Soner boundary condition allows for smooth representations also to reach exact solutions.

\section{Quantifying the Role of Data}

This section presents a systematic evaluation of the role of data losses in the stability of learning geodesic distances.
In two ablation experiments we aim to isolate the effect of varying data quantity and data quality on the convergence of the learning process.
As a key takeaway, the results suggest that even sparse and noisy data can significantly enhance learning.

The first ablation study investigates how the \textit{amount of supervision} affects convergence and accuracy.
A desirable outcome of the ablation is to determine the minimum level of supervision necessary for reliable geodesic estimation and whether additional data continues to yield improvements in convergence robustness.

The second ablation study focuses on how supervision on imperfect data influences learning. In real-world applications, geodesic distance estimates are generally inaccurate and overestimating.
These experiments help determine how robust the learned geodesic representation is to noisy or inaccurate supervision, providing insights into the practical reliability of physics-informed neural motion planning in real-world settings.

\begin{figure}
    \centering
    \includegraphics[width=\linewidth]{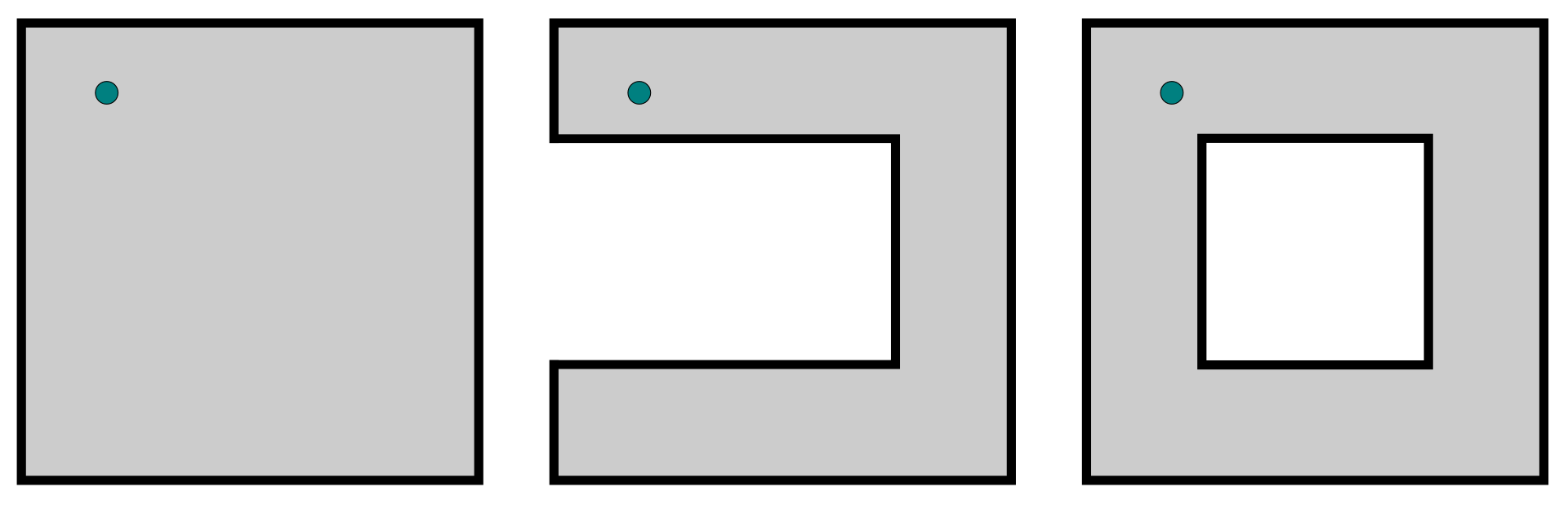}
    \caption{Canonical spaces designed for the ablation study on the effect of data losses on learning. From left to right: a convex domain, a non-convex domain, a non-simply connected domain. The shadowed area represents the free space, and the boundary is represented by the black lines. An example source point in the domain is shown in blue.}
    \label{fig:canonical_spaces}
\end{figure}

\subsection{Experimental Setup}

We evaluate our method on three canonical 2D domains illustrated in Figure~\ref{fig:canonical_spaces}: convex, non-convex, and non-simply-connected. 
For each domain and trial, we approximate the geodesic distance to a goal or source point using a Residual MLP (ResMLP) architecture (5 layers with 128 hidden units per layer, i.e., $5 \times 128$) as described in Section~\ref{sec:framework}.
Experiments were computed on a system with a \texttt{i7-13800H} processor without GPU acceleration.

Unless specified, the loss function is composed of the three components: data loss, physics loss, and boundary loss, as presented in Section~\ref{sec:framework}. The weighting for these three losses is held fixed (equal weight) across all scenarios.

A batch size ($256$) is used for the interior and boundary samples, while for the data component the batch size is set to the size of the dataset if it is smaller than the nominal value. 
Sampling is performed uniformly: in the interior via rejection sampling; on the boundary by mapping the unit interval onto the boundary.

Due to the possibility that some environments may lead the optimization to converge to suboptimal solutions, we define a convergence criterion that combines (i) a minimum number of iterations to avoid premature termination (1000), and (ii) a plateau in the total loss measured over a sliding window, set to an average loss difference over $100$ iterations of $10^{-4}$.

\subsection{Data Quantity Ablation}

We compare a range of training scenarios with varying intermediate levels of supervision, defined as the number of points in a labeled dataset containing the corresponding distances and gradients.
 
We consider the following six cases: physics-only (no data), and datasets of sizes~${[1, 10, 100, 1000,\infty]}$.
In the infinite data scenario, physics losses are not active, and batches are sampled at each iteration directly from the domain.

For each domain and data scenario, we perform $12$ independent trials with different random seeds and source points. 
The source points are uniformly sampled from the upper left region of the domain for more reliable statistics over few trials.
For each trial we compute the maximum absolute error of the distance, and gradient approximations.
On the boundary of the domain we evaluate the maximum violation of the Soner condition.
The results are shown in Table~\ref{tab:error_stats}, via the mean, standard deviation, minimum, and maximum of the maximum absolute errors over $12$ trials. The distance error for the environments is visualized in Figure~\ref{fig:data-error}.

\subsubsection{Results and Analysis}

\renewcommand{\arraystretch}{1.2} 

\begin{table*}[ht]
    \centering
    \small 
    \setlength{\tabcolsep}{8pt} 
    \caption{Maximum absolute error statistics over $12$ trials. Mean ± Std in the first row, and (Min, Max) in the second row.}
    \label{tab:error_stats}
    \begin{tabular}{c ccc ccc}
        \toprule
        \multirow{3}{*}{Dataset Size} & \multicolumn{3}{c}{Non-Convex Domain} & \multicolumn{3}{c}{Non-simply Connected Domain} \\
        \cmidrule(lr){2-4} \cmidrule(lr){5-7}
        & $e_\mathrm{distance}$ & $e_\mathrm{gradient}$ & $e_\mathrm{boundary}$ & $e_\mathrm{distance}$ & $e_\mathrm{gradient}$ & $e_\mathrm{boundary}$ \\
        \midrule
        0   & 2.23 ± 0.39  & 2.06 ± 0.16  & 0.236 ± 0.247  & 0.483 ± 0.102  & 1.82 ± 0.27  & 0.264 ± 0.315  \\
            & (1.31, 2.73)  & (1.93, 2.48)  & (0.00, 0.70)  & (0.33, 0.65)  & (1.39, 2.15)  & (0.00, 0.88)  \\
        \midrule
        1   & 2.19 ± 0.74  & 1.94 ± 0.30  & 0.611 ± 0.381  & 0.355 ± 0.194  & 1.44 ± 0.38  & 0.521 ± 0.299  \\
            & (0.09, 2.72)  & (1.38, 2.50)  & (0.05, 1.00)  & (0.05, 0.64)  & (0.93, 2.12)  & (0.25, 1.00)  \\
        \midrule
        10  & 0.565 ± 0.894  & 1.59 ± 0.45  & 0.375 ± 0.336  & 0.196 ± 0.144  & 1.35 ± 0.33  & 0.501 ± 0.320  \\
            & (0.05, 2.22)  & (1.04, 2.40)  & (0.01, 0.98)  & (0.05, 0.56)  & (0.86, 1.96)  & (0.00, 0.98)  \\
        \midrule
        100 & 0.057 ± 0.018  & 1.37 ± 0.52  & 0.510 ± 0.324  & 0.164 ± 0.088  & 1.27 ± 0.27  & 0.738 ± 0.274  \\
            & (0.03, 0.08)  & (0.65, 2.45)  & (0.01, 1.01)  & (0.05, 0.31)  & (0.92, 1.81)  & (0.24, 1.04)  \\
        \midrule
        1000 & 0.051 ± 0.022  & 1.01 ± 0.23  & 0.627 ± 0.252  & 0.146 ± 0.090  & 1.36 ± 0.40  & 0.585 ± 0.279  \\
             & (0.02, 0.10)  & (0.73, 1.41)  & (0.22, 0.99)  & (0.03, 0.31)  & (0.82, 2.08)  & (0.18, 1.06)  \\
        \midrule
        $\infty$ & 0.027 ± 0.013  & 0.788 ± 0.129  & 0.964 ± 0.029  & 0.127 ± 0.089  & 0.903 ± 0.113  & 0.944 ± 0.104  \\
                 & (0.01, 0.06)  & (0.53, 1.03)  & (0.90, 1.00)  & (0.02, 0.31)  & (0.73, 1.11)  & (0.67, 1.09)  \\
        \bottomrule
    \end{tabular}
\end{table*}

All models converged near the lower bound of iterations ($1000$), with mean and standard deviation number of iterations at $1053 \pm 71$.
Since the loss convergence criteria were reached, this indicates that a conservative lower bound of iterations was chosen.

In the convex environment, the model consistently converged across all dataset sizes, achieving mean maximum distance errors below \(10^{-3}\). 
A general trend of increased accuracy with larger dataset sizes is observed, with decreasing error values in the order of $10^{-4}$. 
Additionally, the Soner condition was exactly satisfied in all trials, confirming that viscosity solutions of the Eikonal were well approximated in this setting.
For the other two environments, results are presented in Table~\ref{tab:error_stats} and discussed below.

Across the non-convex and non-simply connected environments, distance and gradient errors tend to decrease with more data in all presented statistics. 
However, a statistic that stands out is the standard deviation of distance errors:
\begin{itemize}
    \item Small datasets ($1$ and $10$ points) exhibit much higher error variance compared to larger datasets.
    \item The effect is particularly strong in the non-convex environment (more than $10\times$ increase) and also present in the third environment ($2\times$ increase).
    \item Despite this, the minimum error remains in the same order of magnitude across all dataset sizes, except in the physics-only case.
\end{itemize}

\begin{figure}
    \centering
    \includegraphics[width=\linewidth]{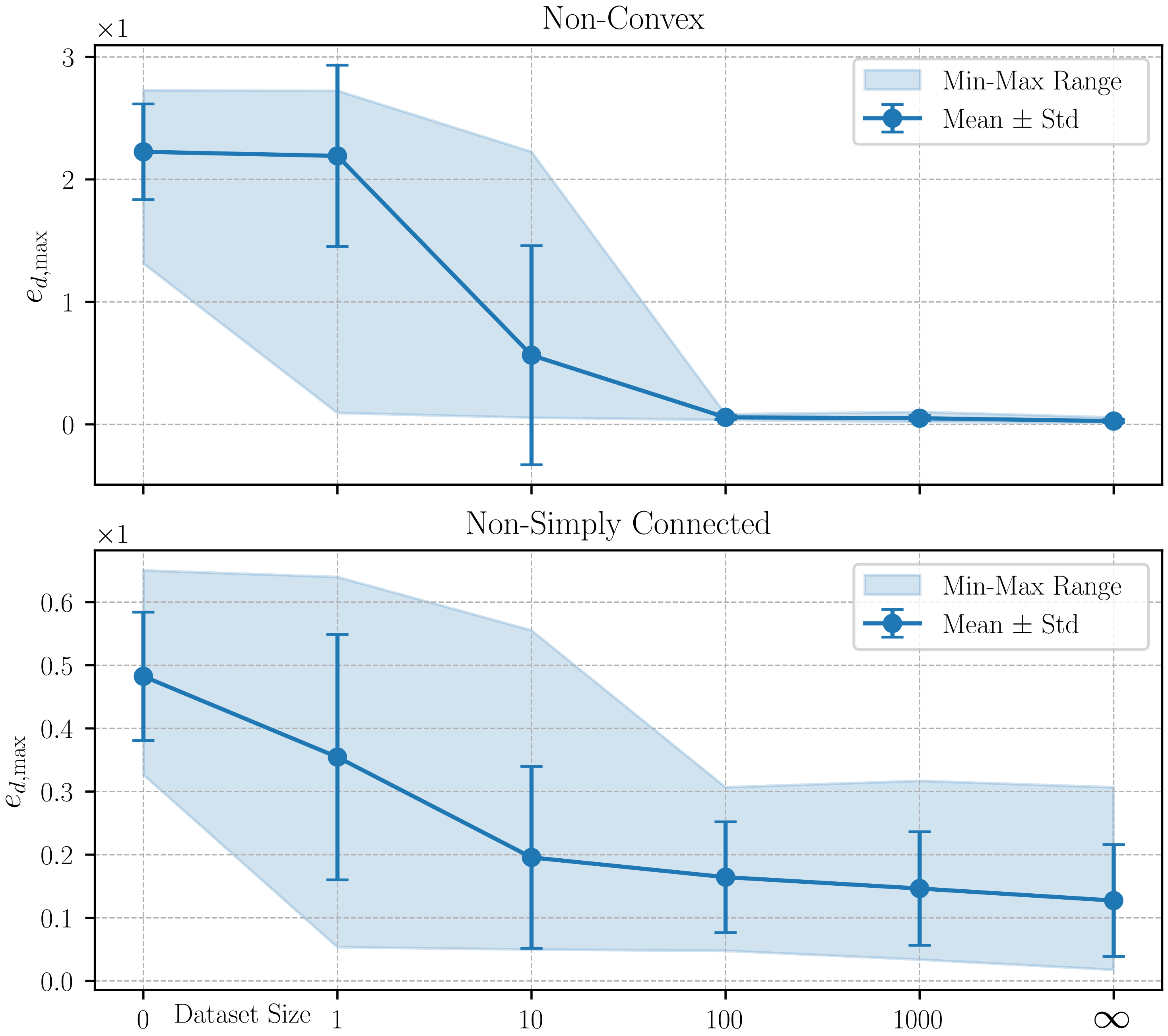}
    \caption{The maximum absolute distance error at different dataset sizes are visualized in this figure. The statistics are shown for the non-convex (top) and the non-simply connected (bottom) environments over $12$ different trials. A general trend is observed of declining mean error with an increasing dataset size. The standard deviation of the errors at dataset sizes of $1$ and $10$ are considerably higher than at larger datasets.}
    \label{fig:data-error}
\end{figure}

These observations are visualized in Figure~\ref{fig:data-error}, and suggest that the specific choice of sampled points matters more than their sheer quantity. 
This thesis is further supported by Figure~\ref{fig:one-matters}, which demonstrates how a single well-placed sample can significantly guide the approximation toward lower error regions. 
Additionally, this highlights the regularizing effect of physics-based losses, which help extrapolate sparse data more effectively across the domain.

No clear trend emerges regarding boundary condition violations as a function of dataset size. 
For instance, the smallest boundary violations (among non-empty datasets) occurred at size $10$. 
Models trained with fully supervised distances and gradients exhibited larger boundary condition violations than those incorporating active boundary losses in smaller datasets. 
This further suggests that physics-losses can improve solutions, even in cases where regression targets are available and abundant.

\subsection{Data Quality Ablation}

The second ablation study focuses on how imperfect data influences learning. To simulate imperfect data, we introduce controlled noise into the training data and assess its effect on model performance. We use the following noise models with a shared noise level $\eta$.

Given a geodesic distance $d$, we apply multiplicative noise sampled from an exponential distribution
\begin{align}
    d' = d + d \xi, \quad \xi \sim \mathrm{Exp}(\lambda), \quad \lambda = \frac{1}{\eta}.
\end{align}
This simulates inexact shortest path estimates where noise magnitude is assumed to be proportional to distance.    

Given a gradient $g$ of the geodesic distance, we introduce additive Gaussian noise
\begin{align}
    g' = \frac{g + \nu}{\| g + \nu \|},\quad \nu \sim \mathcal{N}(0, \sigma^2 I),
\end{align}
where $\mathcal{N}(0, \sigma^2 I)$ represents isotropic Gaussian noise with standard deviation $\sigma = \eta$. The result is normalized to preserve gradients of unit norm.

The evaluation is performed exclusively in the non-convex environment and for dataset sizes $[1, 10]$.
A total of $24$ datasets are generated, of which $12$ contain $10$ samples, and $12$ contain $1$ sample.
Each dataset is corrupted according to the noise levels ${\eta = [0, 0.05, 0.10, 0.25]}$, simulating a wide range of plausible inaccuracies.
These choices were motivated by the high variance of distance errors at these configurations in the data quantity ablation, thus, we focus on the distance error statistics.
The results are summarized in Table~\ref{tab:noise_stats} and discussed below.

\subsubsection{Results and Analysis}

\begin{figure}
    \centering
    \includegraphics[width=\linewidth]{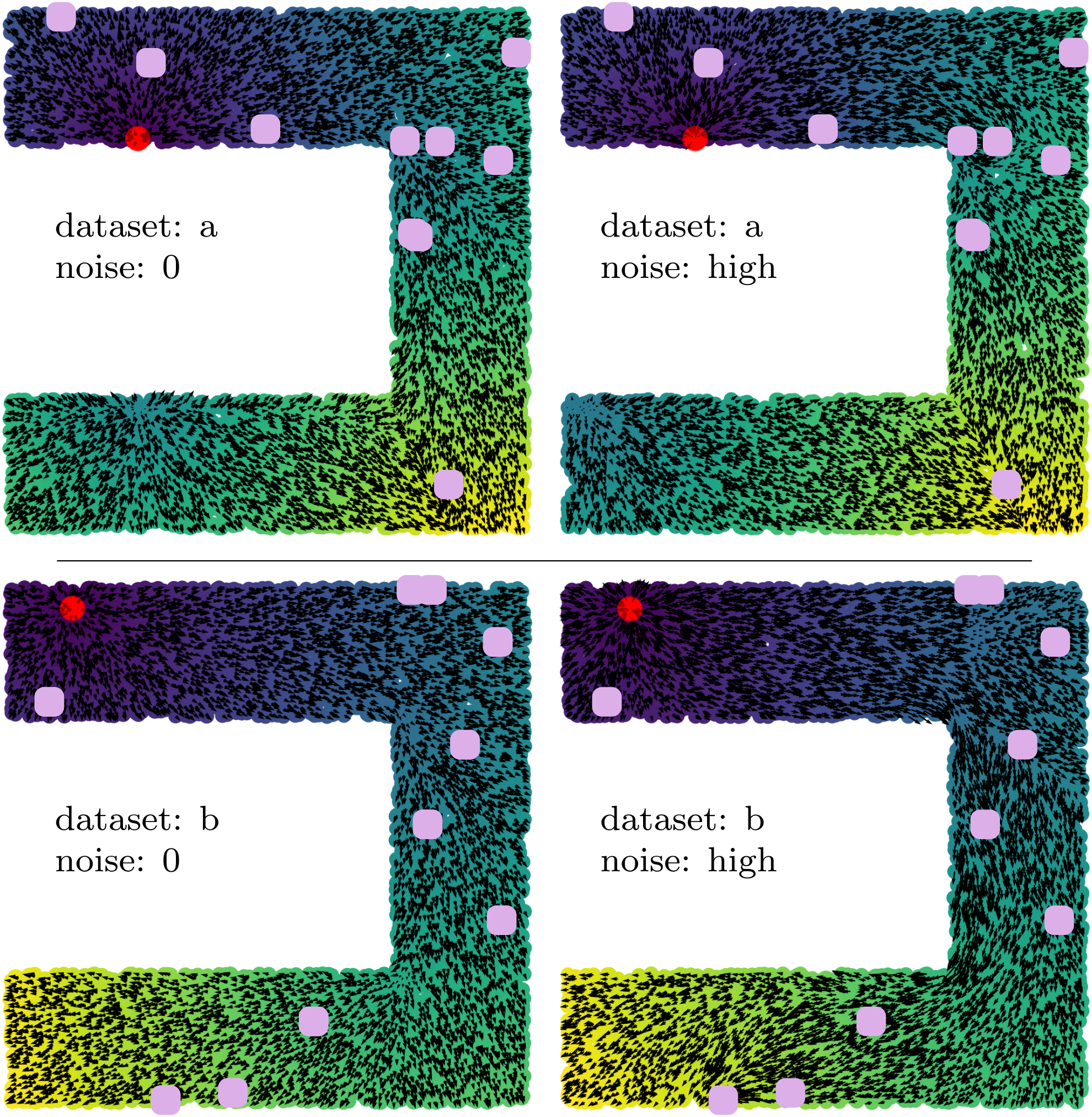}
    \caption{Visualization of the robustness to noise, using two different datasets with a fixed number of points ($10$), and the two extreme levels of noise~${\eta=[0, 0.25]}$. The source point is depicted in red; the points in the dataset are shown in light purple. Both trials converge to qualitatively similar solutions at all evaluated noise levels. The dataset (a) guides convergence to a poor minimum, while dataset (b) guides solutions to good approximations.}
    \label{fig:noise-robust}
\end{figure}

\renewcommand{\arraystretch}{1.2} 

\begin{table}[ht]
    \centering
    \small 
    \setlength{\tabcolsep}{8pt} 
    \caption{Maximum absolute error statistics ($e_\mathrm{distance}$) for the non-convex environment, over $12$ trials. Mean ± Std in the first row, and (Min, Max) in the second row.}
    \label{tab:noise_stats}

    \begin{tabular}{c cc}
        \toprule
        \multirow{2}{*}{Noise Level} & \multicolumn{2}{c}{Dataset Size} \\
        \cmidrule(lr){2-3}
        & 1 & 10 \\
        \midrule
        0.00  & 2.191 ± 0.740  & 0.565 ± 0.894  \\
              & (0.092, 2.723)  & (0.054, 2.225)  \\
        \midrule
        0.05  & 2.178 ± 0.656  & 0.732 ± 0.888  \\
              & (0.227, 2.735)  & (0.085, 2.318)  \\
        \midrule
        0.10  & 2.192 ± 0.640  & 0.849 ± 0.850  \\
              & (0.348, 2.853)  & (0.070, 2.362)  \\
        \midrule
        0.25  & 2.198 ± 0.558  & 1.242 ± 0.717  \\
              & (0.722, 2.728)  & (0.286, 2.300)  \\
        \bottomrule
    \end{tabular}
\end{table}

For datasets of size $1$, the mean error remains nearly constant across noise levels, suggesting that adding noise does not significantly impact accuracy. 
In contrast, for datasets of size $10$, the mean error increases with noise, indicating a growing sensitivity to data corruption.

For both dataset sizes, the standard deviation of the error remains stable, showing no clear dependency on noise. 
However, the minimum error tends to increase with noise, particularly at the highest noise level, while this trend is less pronounced for lower noise levels. 
Overall, while the effect is weak, the expected relationship holds, i.e., more noise leads to higher errors, though the observed robustness suggests that extreme degradation is avoided at low noise levels.

A key result is the remarkable consistency of solutions across noise levels when the underlying dataset is fixed. 
The mean standard deviation of the errors across noise levels is $0.22$, more than three times smaller (!) than the same statistic computed across different datasets ($0.74$). 
This suggests that, given a fixed uncorrupted dataset, the learned geodesic approximation converges to similar solutions, regardless of added noise (within reasonable bounds).
This robustness is illustrated in Figure~\ref{fig:noise-robust}, which compares solutions obtained from two different datasets under varying noise levels.

\section{Discussion}

Sparse, inaccurate data can have a large positive effect on convergence and reliability in learning geodesic distances. Our experiments support this claim, but its validity must be considered within the context of the evaluation framework and its inherent limitations.

\subsection{Architectural Considerations}

The network architecture (ResMLP with $5 \times 128$ configuration) was chosen based on its demonstrated ability to converge in the data-only regime. 
While capable of representing the true solution, optimization challenges in the PDE-only case may prevent convergence to the same solution. 
Due to computational and space constraints, we fix the architecture in this study and focus exclusively on the effect of data losses.

\subsection{Physics-only Baseline and Scalability}

The performance of physics-only training was highly sensitive to initialization, as no additional convergence heuristics or adaptive sampling strategies were employed. 
This choice was deliberate, as we aim to explore methods that remain feasible in higher dimensions, where common variance reduction techniques become computationally intractable~\cite{cuomo2022scientific}.

While our experiments focus on static 2D domains, the method is conceptually extendable to higher-dimensional configuration spaces, and to dynamic environments, based on the success of related methods~\cite{ni2022ntfields, li2024riemannian, fishman2023motion}.

\subsection{Sparse but Well-Positioned Data}

In high-dimensional and complex environments, identifying key data points remains a challenge. 
A pessimistic interpretation of our findings suggests that extremely large datasets may be required for convergence if the likelihood of encountering well-posed samples is low. 
However, structured demonstrations and first-principles reasoning can mitigate this issue, providing targeted supervision rather than relying on random sampling.
This motivates future work on learning to predict distributions of well-positioned points in unseen environments.  

\subsection{Soft Boundary Constraints}

If perfectly enforced, soft boundary constraints ensure that no path following the negative gradient leaves the domain. However, they are generally not guaranteed to be satisfied. When applied directly to planning methods such as NTFields, a conservative bias can be introduced by imposing $n \cdot \nabla d \leq \alpha$ for some $\alpha \in [0, -1]$. Additional verification techniques, such as conformal prediction \cite{tayal2025physicsinformedmachinelearningframework}, may offer further robustness.

\section{Conclusion}
This work introduces a framework for learning geodesic distances in bounded domains using the Soner boundary condition and investigates how data losses impact the stability and accuracy of neural geodesic distance solvers.
By augmenting physics-informed training with sparse supervision, we demonstrate that a small number of well-placed data points (\(1\)–\(10\)) is sufficient to match the accuracy of fully supervised approaches while improving adherence to boundary constraints. 
These findings suggest that hybrid physics-data approaches can significantly enhance convergence reliability in learning-based geodesic distance estimation, and therefore effectively be applied for physics-informed neural motion planning.

\addtolength{\textheight}{-12cm}   








\balance
\bibliographystyle{IEEEtran}
\bibliography{ref}

\begin{thebibliography}{10}
\providecommand{\url}[1]{#1}
\csname url@samestyle\endcsname
\providecommand{\newblock}{\relax}
\providecommand{\bibinfo}[2]{#2}
\providecommand{\BIBentrySTDinterwordspacing}{\spaceskip=0pt\relax}
\providecommand{\BIBentryALTinterwordstretchfactor}{4}
\providecommand{\BIBentryALTinterwordspacing}{\spaceskip=\fontdimen2\font plus
\BIBentryALTinterwordstretchfactor\fontdimen3\font minus \fontdimen4\font\relax}
\providecommand{\BIBforeignlanguage}[2]{{%
\expandafter\ifx\csname l@#1\endcsname\relax
\typeout{** WARNING: IEEEtran.bst: No hyphenation pattern has been}%
\typeout{** loaded for the language `#1'. Using the pattern for}%
\typeout{** the default language instead.}%
\else
\language=\csname l@#1\endcsname
\fi
#2}}
\providecommand{\BIBdecl}{\relax}
\BIBdecl

\bibitem{sethian1996fast}
J.~A. Sethian, ``A fast marching level set method for monotonically advancing fronts.'' \emph{proceedings of the National Academy of Sciences}, vol.~93, no.~4, pp. 1591--1595, 1996.

\bibitem{zhao2005fast}
H.~Zhao, ``A fast sweeping method for eikonal equations,'' \emph{Mathematics of computation}, vol.~74, no. 250, pp. 603--627, 2005.

\bibitem{smith2020eikonet}
J.~D. Smith, K.~Azizzadenesheli, and Z.~E. Ross, ``Eikonet: Solving the eikonal equation with deep neural networks,'' \emph{IEEE Transactions on Geoscience and Remote Sensing}, vol.~59, no.~12, pp. 10\,685--10\,696, 2020.

\bibitem{grubas2023neural}
S.~Grubas, A.~Duchkov, and G.~Loginov, ``Neural eikonal solver: Improving accuracy of physics-informed neural networks for solving eikonal equation in case of caustics,'' \emph{Journal of Computational Physics}, vol. 474, p. 111789, 2023.

\bibitem{cuomo2022scientific}
S.~Cuomo, V.~S. Di~Cola, F.~Giampaolo, G.~Rozza, M.~Raissi, and F.~Piccialli, ``Scientific machine learning through physics--informed neural networks: Where we are and what’s next,'' \emph{Journal of Scientific Computing}, vol.~92, no.~3, p.~88, 2022.

\bibitem{deckelnick2011numerical}
K.~Deckelnick, C.~M. Elliott, and V.~Styles, ``Numerical analysis of an inverse problem for the eikonal equation,'' \emph{Numerische Mathematik}, vol. 119, pp. 245--269, 2011.

\bibitem{soner1986optimal}
H.~M. Soner, ``Optimal control with state-space constraint i,'' \emph{SIAM Journal on Control and Optimization}, vol.~24, no.~3, pp. 552--561, 1986.

\bibitem{treister2016fast}
E.~Treister and E.~Haber, ``A fast marching algorithm for the factored eikonal equation,'' \emph{Journal of Computational physics}, vol. 324, pp. 210--225, 2016.

\bibitem{belyaev2015variational}
A.~G. Belyaev and P.-A. Fayolle, ``On variational and pde-based distance function approximations,'' in \emph{Computer Graphics Forum}, vol.~34, no.~8.\hskip 1em plus 0.5em minus 0.4em\relax Wiley Online Library, 2015, pp. 104--118.

\bibitem{crane2013geodesics}
K.~Crane, C.~Weischedel, and M.~Wardetzky, ``Geodesics in heat: A new approach to computing distance based on heat flow,'' \emph{ACM Transactions on Graphics (TOG)}, vol.~32, no.~5, pp. 1--11, 2013.

\bibitem{bin2021pinneik}
U.~bin Waheed, E.~Haghighat, T.~Alkhalifah, C.~Song, and Q.~Hao, ``Pinneik: Eikonal solution using physics-informed neural networks,'' \emph{Computers \& Geosciences}, vol. 155, p. 104833, 2021.

\bibitem{ni2022ntfields}
R.~Ni and A.~H. Qureshi, ``Ntfields: Neural time fields for physics-informed robot motion planning,'' \emph{arXiv preprint arXiv:2210.00120}, 2022.

\bibitem{ni2023progressive}
------, ``Progressive learning for physics-informed neural motion planning,'' \emph{arXiv preprint arXiv:2306.00616}, 2023.

\bibitem{liu2024physics}
Y.~Liu, R.~Ni, and A.~H. Qureshi, ``Physics-informed neural mapping and motion planning in unknown environments,'' \emph{arXiv preprint arXiv:2410.09883}, 2024.

\bibitem{li2024riemannian}
Y.~Li, J.~Qiu, and S.~Calinon, ``A riemannian take on distance fields and geodesic flows in robotics,'' \emph{arXiv preprint arXiv:2412.05197}, 2024.

\bibitem{hahn2024laplacian}
J.~Hahn, K.~Mikula, and P.~Frolkovi{\v{c}}, ``Laplacian regularized eikonal equation with soner boundary condition on polyhedral meshes,'' \emph{Computers \& Mathematics with Applications}, vol. 156, pp. 74--86, 2024.

\bibitem{inproceedings}
Y.~Li, X.~Chi, A.~Razmjoo, and S.~Calinon, ``Configuration space distance fields for manipulation planning,'' 07 2024.

\bibitem{kelshaw2024computing}
D.~Kelshaw and L.~Magri, ``Computing distances and means on manifolds with a metric-constrained eikonal approach,'' \emph{arXiv preprint arXiv:2404.08754}, 2024.

\bibitem{implicitLipman}
A.~Gropp, L.~Yariv, N.~Haim, M.~Atzmon, and Y.~Lipman, ``Implicit geometric regularization for learning shapes,'' in \emph{Proceedings of the 37th International Conference on Machine Learning}, ser. ICML'20.\hskip 1em plus 0.5em minus 0.4em\relax JMLR.org, 2020.

\bibitem{fishman2023motion}
A.~Fishman, A.~Murali, C.~Eppner, B.~Peele, B.~Boots, and D.~Fox, ``Motion policy networks,'' in \emph{Conference on Robot Learning}.\hskip 1em plus 0.5em minus 0.4em\relax PMLR, 2023, pp. 967--977.

\bibitem{tayal2025physicsinformedmachinelearningframework}
\BIBentryALTinterwordspacing
M.~Tayal, A.~Singh, S.~Kolathaya, and S.~Bansal, ``A physics-informed machine learning framework for safe and optimal control of autonomous systems,'' 2025. [Online]. Available: \url{https://arxiv.org/abs/2502.11057}
\BIBentrySTDinterwordspacing

\end{thebibliography}

\end{document}